\documentclass{article}
\usepackage{spconf,amsmath,graphicx,subfigure}
\usepackage{gensymb}
\usepackage{booktabs}
\usepackage{url}
\usepackage{siunitx}

\title{End-to-end Audiovisual Speech Recognition\thanks{Accepted to ICASSP 2018}}
\name{\begin{tabular}{c}Stavros Petridis$^1$, Themos Stafylakis$^2$, Pingchuan Ma$^1$, Feipeng Cai$^1$ \\
Georgios Tzimiropoulos$^2$, Maja Pantic$^{1}$\end{tabular}}
\address{$^1$Dept. of Computing, Imperial College London, UK \\
$^2$Computer Vision Laboratory, University of Nottingham, UK \\
stavros.petridis04@imperial.ac.uk, themos.stafylakis@nottingham.ac.uk \\
}
\begin{document}
%
\maketitle
\begin{abstract}
Several end-to-end deep learning approaches have been recently presented which extract either audio or visual features from the input images or audio signals and perform speech recognition. However, research on end-to-end audiovisual models is very limited. In this work, we present an end-to-end audiovisual model based on residual networks and Bidirectional Gated Recurrent Units (BGRUs). To the best of our knowledge, this is the first audiovisual fusion model which simultaneously learns to extract features directly from the image pixels and audio waveforms and performs within-context word recognition on a large publicly available dataset (LRW).  The model consists of two streams, one for each modality, which extract features directly from mouth regions and raw waveforms. The temporal dynamics in each stream/modality are modeled by a 2-layer BGRU and the fusion of multiple streams/modalities takes place via another 2-layer BGRU. A slight improvement in the classification rate over an end-to-end audio-only and MFCC-based model is reported in clean audio conditions and low levels of noise. In presence of high levels of noise, the end-to-end audiovisual model significantly outperforms both audio-only models.
\end{abstract}
\section{Introduction}
\label{sec:intro}

Traditional audiovisual fusion systems consist of two stages, feature extraction from the image and audio signals and combination of the features for joint classification \cite{Potamianos2003,Dupont2000,predBasedAVfusion}. Recently, several deep learning approaches for audiovisual fusion have been presented which aim to replace the feature extraction stage with deep bottleneck architectures. Usually a transform, like principal component analysis (PCA), is first applied to the mouth region of interest (ROI) and spectrograms or concatenated Mel-Frequency Cepstral Coefficients (MFCCs) and a deep autoencoder is trained to extract bottleneck features \cite{ngiam2011multimodal,hu2016temporal,ninomiya2015integration,mroueh2015,takashima2016audio,petridisPantic_icassp2016}. Then these features are fed to a classifier like a support vector machine or a Hidden Markov Model. 

Few works have been presented very recently which follow an end-to-end approach for visual speech recognition. The main approaches followed can be divided into two groups. In the first one, fully connected layers are used to extract features and LSTM layers model the temporal dynamics of the sequence \cite{petridis2017deepVisualSpeech,wand2016lipreading}. In the second group, a 3D convolutional layer is used followed either by standard convolutional layers \cite{assael2016lipnet} or residual networks (ResNet) \cite{stafylakis2017combining} combined with LSTMs or GRUs. End-to-end approaches have also been successfully used for speech emotion recognition using 1D CNNs and LSTMs \cite{trigeorgis2016adieu}.

However, work on end-to-end audiovisual speech recognition has been very limited. To the best of our knowledge, there are only two works which perform end-to-end training for audiovisual speech recognition \cite{chung2016lipSentences,end2endAV}. In the former, an attention mechanism is applied to both the mouth ROIs and MFCCs and the model is trained end-to-end. However, the system does not use the raw audio signal or spectrogram but relies on  MFCC features. In the latter, fully connected layers together with LSTMs are used in order to extract features directly from raw images and spectrograms and perform classification on the OuluVS database \cite{Anina2015}.

In this paper, we extend the work of \cite{petridis2017deepVisualSpeech}, which mainly works for small databases, using ResNets as proposed in \cite{stafylakis2017combining}. To the best of our knowledge, this is the first end-to-end model which performs audiovisual word recognition from raw mouth ROIs and waveforms on a large in-the-wild database. The proposed model consists of two streams, one per modality,  which extract features directly from the raw images and waveforms, respectively. Each stream consists of a ResNet which extracts features from the raw inputs. This is followed by a 2-layer BGRU network which models the temporal dynamics in each stream. Finally,  the information of the different streams/modalities is fused  via another 2-layer BGRU which models the joint temporal dynamics. A similar architecture has been proposed by \cite{endtoendtzirakis2017} for audiovisual emotion recognition. The main differences of our work are the following: 1) we use a ResNet for the audio stream instead of a rather shallow 2-layer CNN, 2) we do not use a pre-trained ResNet for the visual stream but we train a ResNet from scratch, 3) we use BGRUs in each stream which help modeling the temporal dynamics of each modality instead of using just one BLSM layer at the top and  4) we use a training procedure which allows for efficient end-to-end training of the entire network.

We perform classification of 500 words from the LRW database achieving state-of-the-art performance for audiovisual fusion. The proposed system  results in an absolute increase of 0.3\% in classification accuracy over the end-to-end audio-only model and an MFCC-based system. The end-to-end audiovisual fusion model also significantly outperforms (up to 14.1\% absolute improvement) the audio-only models under high levels of noise.

\begin{figure}[t]

\begin{minipage}[t]{0.75\linewidth}
  \centering
  \subfigure[]{\includegraphics[width=0.45\linewidth, trim=15 15 15 15, clip]{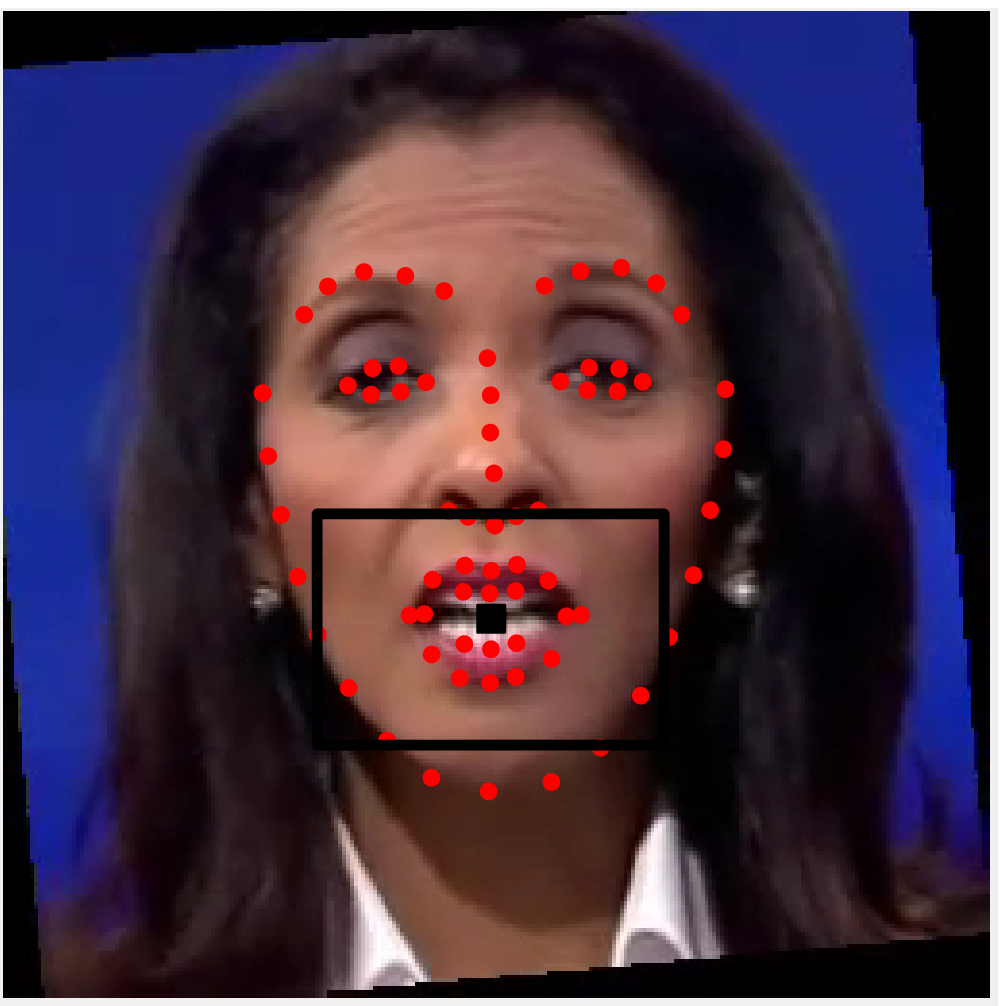}}  
  \subfigure[]{\includegraphics[width=0.45\linewidth, trim=5 5 5 5, clip] {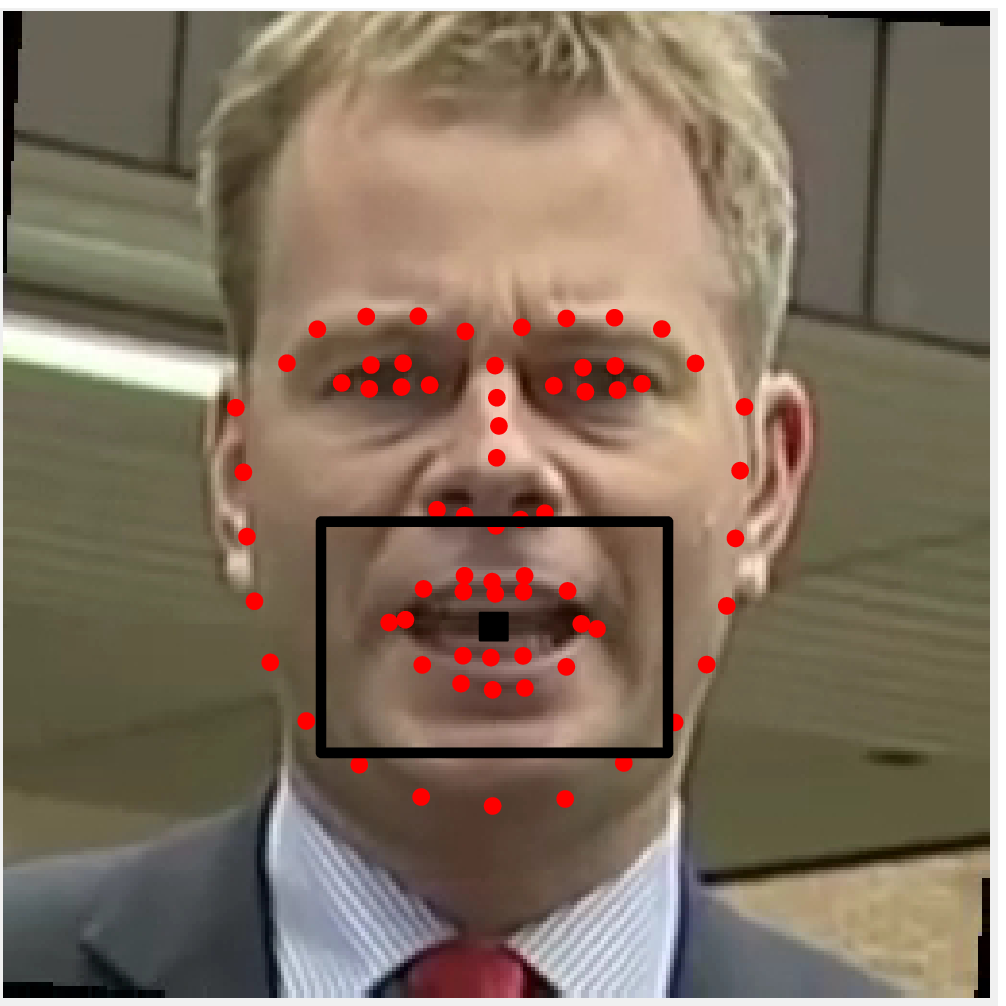}}   

\caption{Example of mouth ROI extraction.}
\label{fig:mouthROI}
\end{minipage}

\end{figure}

\section{LRW Database}
\label{sec:database}

For the purposes of this study we use the Lip Reading in the Wild (LRW) database \cite{chung2016lip} which is the largest publicly available lipreading dataset in the wild. The database consists of short segments (1.16 seconds) from BBC programs, mainly news and talk shows. It is a very challenging set since it contains more than 1000 speakers and large variation in head pose and illumination. The number of words, 500, is also much higher than existing lipreading databases used for word recognition, which typically contain 10 to 50 words \cite{cooke2006,Patterson:2002,Anina2015}. 

Another characteristic of the database is the presence of several words which are visually similar. For example, there are words which are present in their singular and plural forms or simply different forms of the same word, e.g., America and American. We should also emphasise that words appear in the middle of an utterance  and there may be co-articulation of the lips from preceding and subsequent words. 

\begin{figure}[t]

  \centering
\includegraphics[width=0.55\linewidth]{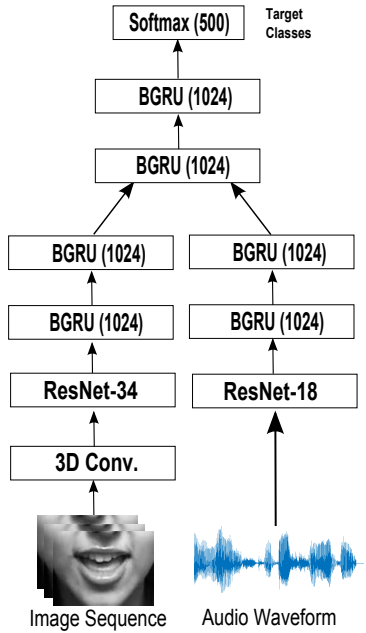}

\caption{Overview of the end-to-end audiovisual speech recognition system. Two streams are used for feature extraction directly from the raw images and audio waveforms.   The temporal dynamics are modelled by BGRUs in each stream. The top two BGRUs fuse the information of the audio and visual streams and jointly model their temporal dynamics. }
\label{fig:system}
\end{figure}

\section{End-to-end Audiovisual Speech Recognition}
\looseness - 1
The proposed deep learning system for audiovisual fusion is shown  in Fig. \ref{fig:system}. It consists of two streams which extract features directly from the raw input images and the audio waveforms, respectively. Each stream consists of two parts: a residual network (ResNet) \cite{he2016deep} which learns to automatically extract features from the raw image and waveform, respectively and a 2-layer BGRU which models the temporal dynamics of the features in each stream. Finally, 2 BGRU layers on top of the two streams are used in order to fuse the information of the audio and visual streams.

\subsection{Visual Stream}
The visual stream is similar to \cite{stafylakis2017combining} and consists of a spatiotemporal convolution 
followed by a 34-layer ResNet and a 2-layer BGRU. A spatiotemporal convolutional layer
is capable of capturing the short-term dynamics of the
mouth region and is proven to be advantageous, even when
recurrent networks are deployed for back-end \cite{assael2016lipnet}. It consists
of a convolutional layer with 64 3D kernels
of 5 by 7 by 7 size (time/width/height), followed by batch normalization and rectified linear units. 

 We use the 34-layer identity mapping
version, which was proposed for ImageNet \cite{he2016identity}. The ResNet drops progressively the
spatial dimensionality until its output
becomes a single dimensional tensor per time step. We should
emphasize that we did not make use of pretrained models, as
they are optimized for completely different tasks (e.g. static
colored images from ImageNet or CIFAR). Finally, the output of ResNet-34 is fed to a 2-layer BGRU which consists of 1024 cells in each layer.

\subsection{Audio Stream}

The audio stream consists of an 18-layer ResNet followed by two BGRU layers. There is no need to use a spatiotemporal convolution front-end in this case as the audio waveform is an 1D signal. We use the standard architecture for the ResNet-18 with the main difference being that we use 1D  instead of  2D kernels which are used for image data. A temporal kernel of 5ms with a stride of 0.25ms is used in the first convolutional layer in order to extract fine-scale spectral information. The output of the ResNet is divided into 29 frames/windows using average pooling in order to ensure the same frame rate as the video is used. These audio frames are then fed to the following ResNet layers which consist of the default kernels of size 3 by 1 so deeper layers extract long-term speech characteristics. The output of the ResNet-18 is fed to a 2-layer BGRU which consists of 1024 cells in each layer (using the same architecture as in \cite{stafylakis2017combining}).

\subsection{Classification Layers}

The BGRU outputs of each stream are concatenated and fed to another 2-layer BGRU in order to fuse the information from the audio and visual streams and jointly model their temporal dynamics. The output layer is a softmax layer which provides a label to each frame. The sequence is labeled based on the highest average probability.

\section{EXPERIMENTAL SETUP}

\subsection{Preprocessing}
\textbf{Video:} The first step is the extraction of the mouth region of interest (ROI). Since the mouth ROIs are already centered, a fixed bounding box of 96 by 96 is used for all videos as shown in Fig. \ref{fig:mouthROI}.  Finally, the frames are transformed to grayscale and are normalized with respect to the overall mean and variance.

\noindent
\textbf{Audio:} Each audio segment is z-normalised, i.e., has zero
mean and standard deviation one to account for variations in different
levels of loudness between the speakers.

\subsection{Evaluation Protocol}

The video segments are already partitioned into training, validation and test sets. There are between 800 and 1000 sequences for each word in the training set and 50 sequences in the validation and test sets, respectively. In total there are 488766, 25000, and 25000 examples in the training, validation and test sets, respectively.

\subsection{Training}
\label{ssec:training}

Training is divided into 2 phases: first the audio/visual streams are trained independently and then the audiovisual network is trained end-to-end. During training data augmentation is performed on the video sequences of mouth ROIs. This is done by applying random cropping and horizontal flips with probability 50\% to all frames of a given clip. Data augmentation is also applied to the audio sequences. During training babble noise at different levels (between -5 dB to 20 db) might be added to the original audio clip. The selection of one of the noise levels or the use of the clean audio is done using a uniform distribution.

\subsubsection{Single Stream Training} 

\textbf{Initialisation:} First, each stream is trained independently. Directly
training end-to-end each stream leads to suboptimal performance so we follow the same 3-step procedure as in \cite{stafylakis2017combining}.
Initially, a temporal convolutional back-end is used instead of the 2-layer BGRU. The combination of ResNet and temporal convolution (together with a softmax output layer) is trained until there is no improvement in the classification rate on the validation set for more than 5 epochs. Then the temporal convolutional back-end is removed and the BGRU back-end is
attached. The 2-layer BGRU (again with a sotfmax output layer) is trained for 5 epochs, keeping the weights of the 3D convolution front-end and the ResNet fixed. 

\looseness - 1
\noindent
\textbf{End-to-End Training:} Once the ResNet and the 2-layer BGRU in each stream have been pretrained then they are put together and the entire stream is trained end-to-end (using a softmax output layer).
The Adam training algorithm \cite{kingma2014adam} is used for end-to-end training with a mini-batch size of 36 sequences and an initial learning rate of 0.0003. Early stopping with a delay of 5 epochs was also used.

\subsubsection{Audiovisual Training} 

\textbf{Initialisation:} Once the single streams have been trained then they are used for initialising the corresponding streams in the multi-stream architecture. Then another 2-layer BGRU is added on top of all streams in order to fuse the single stream outputs. The top BGRU is first trained for 5 epochs (with a softmax output layer), keeping the weights of the audio and visual streams fixed.

\noindent
\textbf{End-to-End Training:} Finally, the entire audiovisual network is trained jointly using Adam with a mini-batch size of 18 sequences and an initial learning rate of 0.0001. Early stopping is also applied with a delay of 5 epochs.

\section{Results}

\begin{table}[t]
\renewcommand{\arraystretch}{1.1}
\renewcommand{\tabcolsep}{7pt}
\caption{ Classification Rate  (CR) of the  Audio-only  (A), Video-only  (V) and audiovisual models (A + V) on the LRW database. *This is a similar end-to-end model which uses a different mouth ROI, computed based on tracked facial landmarks, in each video. In this work, we use a fixed mouth ROI for all videos.  }
\label{tab:resultsBBC}
\centering
\begin{tabular}{lc}
\toprule  Stream &   CR   \\

\midrule A (End-to-End)  &  97.7     \\
A (MFCC) & 97.7 \\
V (End-to-End) & 82.0\\
V \cite{stafylakis2017combining}* & 83.0\\
V \cite{chung2016lipSentences} & 76.2 \\
V \cite{chung2016lip} & 61.1 \\
A + V (End-to-End)  & 98.0  \\

\bottomrule

\end{tabular} 

\end{table}

\looseness - 1
Results are shown in Table \ref{tab:resultsBBC}. We report the performance of the end-to-end audio-only, visual-only and audiovisual models. For comparison purposes, since there are no previous audio/audiovisual results on the LRW database we also compute the performance of a 2-layer BGRU network trained with MFCC features which are the standard features for acoustic speech recognition. We use 13 coefficients (and their deltas) using a 40ms window and a 10ms step. The network is trained in the same way as the BGRU networks in section \ref{ssec:training} with the only difference that it was trained for longer using early stopping. 

The end-to-end audio system results in a similar performance to MFCCs which is a significant result given that the input to the system is just the raw waveform. However, we should note that the effort required in order to train the end-to-end system is significantly higher than the 2-layer BGRU used with MFCCs. The end-to-end audiovisual system leads to a small improvement over the audio-only models of 0.3\%. This is expected since the contribution of the visual modality is usually marginal in clean audio conditions as reported in previous works as well \cite{Potamianos2003,end2endAV}.

In order to investigate the robustness to audio noise of the
audiovisual fusion approach we run experiments under
varying noise levels. The audio signal for each sequence
is corrupted by additive babble noise from the NOISEX
database \cite{varga1993assessment} so as the SNR varies from -5 dB to 20 dB.

\begin{figure}[t]

  \centering
\includegraphics[width=\linewidth]{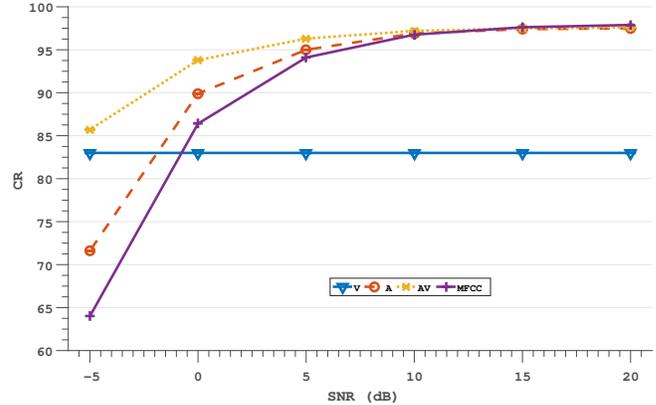}

\caption{ Classification Rate (CR) as a function of the noise level. A: End-to-End audio model. V: End-to-End visual model, AV: End-to-End audiovisual model. MFCC: A 2-layer BGRU trained with MFCCs.}
\label{fig:CRvsNoise}
\end{figure}

Results for the audio, visual and audiovisual models under noisy conditions are shown in Fig.  \ref{fig:CRvsNoise}. The
video-only classifier (blue solid line)
is not affected by the addition of the audio noise and
therefore its performance remains constant over all noise
levels. On the other hand, as expected, the performance
of the audio classifier (red dashed line) is significantly affected. Similarly, the performance of the MFCC classifier (purple solid line) is also significantly affected by noise. It is interesting to point out that although the MFCC and end-to-end audio models result in the same performance when audio is clean or under low levels of noise (10 to 20 dB), the end-to-end audio model results in much better performance under high levels of noise (-5 dB to 5 dB). It results in an absolute improvement of 0.9\%, 3.5\% and 7.5\% over the MFCC classifier, at 5 dB, 0 dB and -5 dB, respectively.  

The audiovisual model (yellow dotted line) 
is more robust to audio noise than the audio-only models. It performs slightly better under low noise levels (10 dB to 20 dB) but it significantly outperforms both of them under high noise levels (-5 dB to 5 dB). In particular, it leads to an absolute improvement of 1.3\%, 3.9\% and 14.1\% over the end-to-end audio-only model at 5 dB, 0 dB and -5 dB, respectively.

\section{Conclusion}
In this work, we present an end-to-end visual audiovisual fusion system which 
jointly learns to extract features directly from the pixels and audio waveforms and performs classification using BGRUs. Results on the largest publicly available database for within-context word recognition in the wild show that the end-to-end audiovisual model slightly outperforms a standard MFCC-based system under clean conditions and low levels of noise. It also significantly outperforms the end-to-end and MFCC-based audio models in the presence of high levels of noise. A natural next step would be to extend the system in order to be able to recognise sentences instead of isolated words. Finally, it would also be interesting to investigate in future work an adaptive fusion mechanism which learns to weight each modality based on the noise levels.

\section{Acknowledgements}
This work has been funded by the European Community Horizon 2020 under grant agreement
no. 645094 (SEWA). Themos Stafylakis has been partly funded by the European Commission program Horizon 2020, under grant agreement no. 706668 (Talking Heads). 


%
%
%




\bibliographystyle{IEEEbib}
\bibliography{mybib}

\end{document}